\documentclass{article}

\usepackage[nonatbib, final]{nips_2018}

\usepackage[utf8]{inputenc} 
\usepackage[T1]{fontenc}    
\usepackage{hyperref}       
\usepackage{url}            
\usepackage{booktabs}       
\usepackage{amsfonts}       
\usepackage{nicefrac}       
\usepackage{microtype}      
\usepackage{graphicx}
\usepackage{color}
\usepackage{placeins}
\usepackage{float}
\usepackage{tabularx,colortbl}
\usepackage{graphics, subfigure, theorem, times, amsfonts, amsmath, amssymb, verbatim}
\usepackage{multicol,multirow,booktabs}
\usepackage{tikz}
\usepackage[linesnumbered,ruled]{algorithm2e}
\usepackage{framed}
\usepackage{authblk}

\usetikzlibrary{shapes,arrows}
\tikzstyle{phantom vertex} = [ ellipse, 
                               anchor = center, 
                               minimum height = 1*\unit, 
                               minimum width  = 1*\unit,
                               inner sep=0pt,
                               anchor=center]
\tikzstyle{red vertex}   = [black, fill = red!20,   phantom vertex, draw]
\tikzstyle{black vertex} = [black, fill = black!20, phantom vertex, draw]
\tikzstyle{blue vertex}  = [black, fill = blue!20,  phantom vertex, draw]
\tikzstyle{green vertex} = [black, fill = green!20,  phantom vertex, draw]
\tikzstyle{yellow vertex} = [black, fill = yellow!20,  phantom vertex, draw]
\tikzstyle{vertex}       = [draw, phantom vertex]

\tikzstyle{point} = [ellipse, inner sep=0pt, draw, fill=white, anchor = center,
                     minimum height = 0.05*\unit, minimum width  = 0.05*\unit]

\usepackage{pgfplots}
\usepackage{color}
\usepackage{needspace}

\input{mysymbol.sty}
\def \QED {$\square$}
\SetKw{KwBy}{by}
\usepackage{setspace}

\title{Modeling Treatment Delays for Patients Using Feature Label Pairs in a Time Series}

\author[1,2]{Weiyu Huang}
\author[1]{Yunlong Wang \thanks{Email: yunlong.wang@iqvia.com}}
\author[1]{Li Zhou}
\author[1]{Emily Zhao}
\author[1]{Yilian Yuan}
\author[2]{Alejandro Ribero}
\affil[1]{Department of Advanced Analytic, IQVIA, Plymouth Meeting, PA}
\affil[2]{Department of Electrical and Computer Engineering, University of Pennsylvania, Philadelphia, PA}

\begin{document}

\maketitle

%
\begin{abstract}
Pharmaceutical targeting is one of key inputs for making sales and marketing strategy planning. Targeting list is built on predicting physician's sales potential of certain type of patient. In this paper, we present a time-sensitive targeting framework leveraging time series model to predict patient's disease and treatment progression. We create time features by extracting service history within a certain period, and record whether the event happens in a look-forward period. Such feature-label pairs are examined across all time periods and all patients to train a model. It keeps the inherent order of services and evaluates features associated to the imminent future, which contribute to improved accuracy.

\end{abstract}

%
\section{Introduction}

A primary goal of oncology pharmaceutical market research is to develop quantitative models for patients that can be used to predict disease progression, e.g., treatment changes of a patient. In most physician alert systems for sales and marketing, the selection of physicians to be included in the alert report is mainly based on their patient’s treatment status. To accurately and quickly identify patient before the treatment decisions even being made would empower foresight ability and competitive edge for marketers, allowing the marketers to focus on the right physicians with the right patients at the right time, before the treatment decisions are made \cite{cravens2006strategic, a2015metaboloepigenetic}. 

Traditionally, physician targeting is achieved by training a regression model to predict a doctor's number of patients that have been diagnosed by certain disease, or the total volume of certain drug this doctor prescribed in a given time window \cite{mak2008rich, hillery2002drug}. Most physician targeting models are trained and applied on doctor level, namely, each sample in the model represents a doctor. For enhancing the targeting accuracy, in this research we propose a time sensitive model to predict doctor's sales potential by forecasting patient's disease progression. Compared with traditional targeting method, by using electronic health records (EHRs) data, the proposed model is built on patient level to predict if a patient will have treatment change in the next given time window. For example, predicting if a cancer patient will start first-line treatment in next three months. Then a doctor's sales potential could be accurately estimated as the number the patients who will start first-line treatment in next three months. Once the messages are deployed, pharma companies may further leverage this data to demonstrate uptake of a treatment or health outcomes as a way to demonstrate effectiveness of the program through return of investment design. 

Recently, machine learning for clinical prediction is growing more popular, and several recent methods have been proposed in predicting disease progression. In \cite{miotto2016deep}, a deep neural network is trained using EHR data to extract patients' latent features that could be used to predict future disease. To address the non-linear temporality in EHR data, a recurrent neural network based model was proposed in \cite{choi2016using} to predict initial diagnosis of heart failure, and compared with conventional methods that ignore temporality. Machine learning algorithms have been proposed to predict patient-level disease progression in areas of Alzheimer's disease\cite{wang2012high}, heart disease\cite{jin2018predicting, gandhi2015predictions}, cancer\cite{singal2013machine}, lung disease\cite{schulam2015framework}, kidney disease\cite{tangri2011predictive}, and the progression of multiple diseases using the Hawkes process\cite{choi2015constructing}. Particularly, machine learning with time series analysis have been applied to process clinical data, for accuracy enhancement\cite{ghassemi2015multivariate, henry2015targeted}.A comprehensive exposition of machine learning with EHR data can be found in  \cite{obermeyer2016predicting, shickel2018deep, ghassemi2018opportunities,ching2018opportunities}. 

In this work, we propose an algorithm combining time series analysis and machine learning, and test it on our real world data-set. The algorithm significantly increase the prediction accuracy comparing with method ignoring time components. Although in this paper we only provide the results of one disease area, this model has also been applied in practical pharmaceutical  problems in Althemer's disease, diabetes and obesity patients' disease progression, and in all of them the deployments are successful which illustrate the robustness of the proposed method. Specifically, the algorithm works as follows. Given the service history for a large number of patients, and the date when a particular event happens to a small group of them, the model aims to predict which patients will receive the event in the near future, e.g. in the next 3 months. The model does so via a novel time-series-based modeling. For each time period of a pre-specified length, the model extracts the service history of a certain patient within the period, and records whether the event happens to the patient in a look-forward period. Such feature-label pairs are generated across all possible time periods and all patients. The model is trained on such generated data using machine-learning-based approach. This model has the advantages of keeping the inherent order of received services, identifying features related to the near future, as well as features yielding no events, all of which are ignored by other algorithms. The proposed algorithm improves accuracy in the imminent future event prediction.

In traditional predictive analysis of pharma marketing research, the objective is identify the predefined event, and trying classify the unknown sample into event and nonevent groups, e.g., \cite{ narayanan2009heterogeneous,kavakiotis2017machine}. All sample information are collected through historical static time period, including event status and sample features. Sample is split into training and testing as two mutually exclusive groups. Model is built with training sample and applied upon testing sample. The validation is performed also on testing sample to measure difference between estimation and observation. Normally these measurements of model accuracy are performed within the same time period. However, when model is applied to sample collected through future time, the validation of model accuracy becomes conversion rate, which is out of estimated event sample, how many will be observed in short time period. This accuracy of model validation through future sample will be significantly reduced in most cases. The time constraint of when event will happen is not supported by traditional predictive modeling. However, this is critical for sales field to plan and measure the trigger alert strategy. The newly developed model is integrating timing component and leveraging forecasting techniques, to re-engineering data structure, event, and feature, and ultimately improve conversion rate in future time frame. The new model shows significant improvement compared with traditional predictive model on conversion rate.

%
\section{Method}

We extract data from IQVIA database including hundreds of millions longitudinal prescription (Rx) and medical claims (Dx). In this study, we focus on one particular cancer, an chronic cancer that there is a watch-and-wait period before patient start first line treatment. IQVIA receives 2 billion prescription claims per year with history from January 2004 with coverage up to 88\% for the retail channel, 50-70\% for traditional and specialty mail order, and 40\% for long-term care. This information represents activities during the prescription transaction including product, provider, payer and geography. In this study, we pulled relevant Rx/Dx data from January 2010 to January 2018.

%
\begin{figure}[t!]
\centerline{

\def \thisplotscale {1}
\def \unit {\thisplotscale cm}

\pgfdeclarelayer{back}
\pgfdeclarelayer{fore}
\pgfdeclarelayer{mid}
\pgfsetlayers{back,mid,fore}

\begin{tikzpicture}[-stealth,  shorten >=0, x = 0.98*\unit, y=0.6*\unit]

\scriptsize
\begin{pgfonlayer}{fore}
    \node at (0, 0) {$w_{1, 1}$};
    \node at (1, 0) {$w_{1, 2}$};
    \node at (2, 0) {$w_{1, 3}$};
    \node at (3, 0) {$w_{1, 4}$};
    \node at (4, 0) {$w_{1, 5}$};
    \node at (5, 0) {$w_{1, 6}$};
    \node at (6, 0) {$w_{1, 7}$};
    \node at (7, 0) {$w_{1, 8}$};
    
    \node at (0, -1) {$w_{2, 1}$};
    \node at (1, -1) {$w_{2, 2}$};
    \node at (2, -1) {$w_{2, 3}$};
    \node at (3, -1) {$w_{2, 4}$};
    \node at (4, -1) {$w_{2, 5}$};
    \node at (5, -1) {$w_{2, 6}$};
    \node at (6, -1) {$w_{2, 7}$};
    \node at (7, -1) {$w_{2, 8}$};
    
    \node at (0, -2) {$w_{3, 1}$};
    \node at (1, -2) {$w_{3, 2}$};
    \node at (2, -2) {$w_{3, 3}$};
    \node at (3, -2) {$w_{3, 4}$};
    \node at (4, -2) {$w_{3, 5}$};
    \node at (5, -2) {$w_{3, 6}$};
    \node at (6, -2) {$w_{3, 7}$};
    \node at (7, -2) {$w_{3, 8}$};
    
    \draw[-] (-3, -2.6) -- (7.5, -2.6);
    \draw[->] (-0.5, -3.7) -- (7.8, -3.7);
    \node at (8.2, -3.7) {time};
    
    \node at (-2, -1) {claim data};
    \node at (-2, -3.2) {labels};
    
    \node at (0, -3.2) {$y_{1}$};
    \node at (1, -3.2) {$y_{2}$};
    \node at (2, -3.2) {$y_{3}$};
    \node at (3, -3.2) {$y_{4}$};
    \node at (4, -3.2) {$y_{5}$};
    \node at (5, -3.2) {$y_{6}$};
    \node at (6, -3.2) {$y_{7}$};
    \node at (7, -3.2) {$y_{8}$};
\end{pgfonlayer}
    
\begin{pgfonlayer}{mid}
    \fill[blue!40!white] (-0.4, 0.4) rectangle (1.4, -0.4);
    \fill[blue!40!white] (-0.4, -0.6) rectangle (1.4, -1.4);
    \fill[blue!40!white] (-0.4, -1.6) rectangle (1.4, -2.4);
    
    \fill[blue!40!white] (1.6, 0.4) rectangle (2.6, -0.4);
    \fill[blue!40!white] (1.6, -0.6) rectangle (2.6, -1.4);
    \fill[blue!40!white] (1.6, -1.6) rectangle (2.6, -2.4);
    
    \fill[blue!40!white] (3.6, -2.8) rectangle (4.4, -3.6);
    
    \fill[blue!40!green] (2.6, 0.4) rectangle (3.4, -0.4);
    \fill[blue!40!green] (2.6, -0.6) rectangle (3.4, -1.4);
    \fill[blue!40!green] (2.6, -1.6) rectangle (3.4, -2.4);
    
    \fill[green!40!white] (3.4, 0.4) rectangle (4.4, -0.4);
    \fill[green!40!white] (3.4, -0.6) rectangle (4.4, -1.4);
    \fill[green!40!white] (3.4, -1.6) rectangle (4.4, -2.4);
    
    \fill[green!40!white] (4.6, 0.4) rectangle (6.4, -0.4);
    \fill[green!40!white] (4.6, -0.6) rectangle (6.4, -1.4);
    \fill[green!40!white] (4.6, -1.6) rectangle (6.4, -2.4);
    
    \fill[green!40!white] (6.6, -2.8) rectangle (7.4, -3.6);
\end{pgfonlayer}
     	
\end{tikzpicture} }
\caption{Generating time-dependent claim-level features where each time index represents one day. For a pre-determined index date $t$, service records for $\delta\Delta$ days prior to the index date are aggregated into $\Delta$ buckets, each with length $\delta$. \vspace{-1mm}}
\label{fig_illustration}

\begin{center}
\includegraphics[trim=1.5cm 0cm 2cm 0cm, clip=false, width=0.7\textwidth]{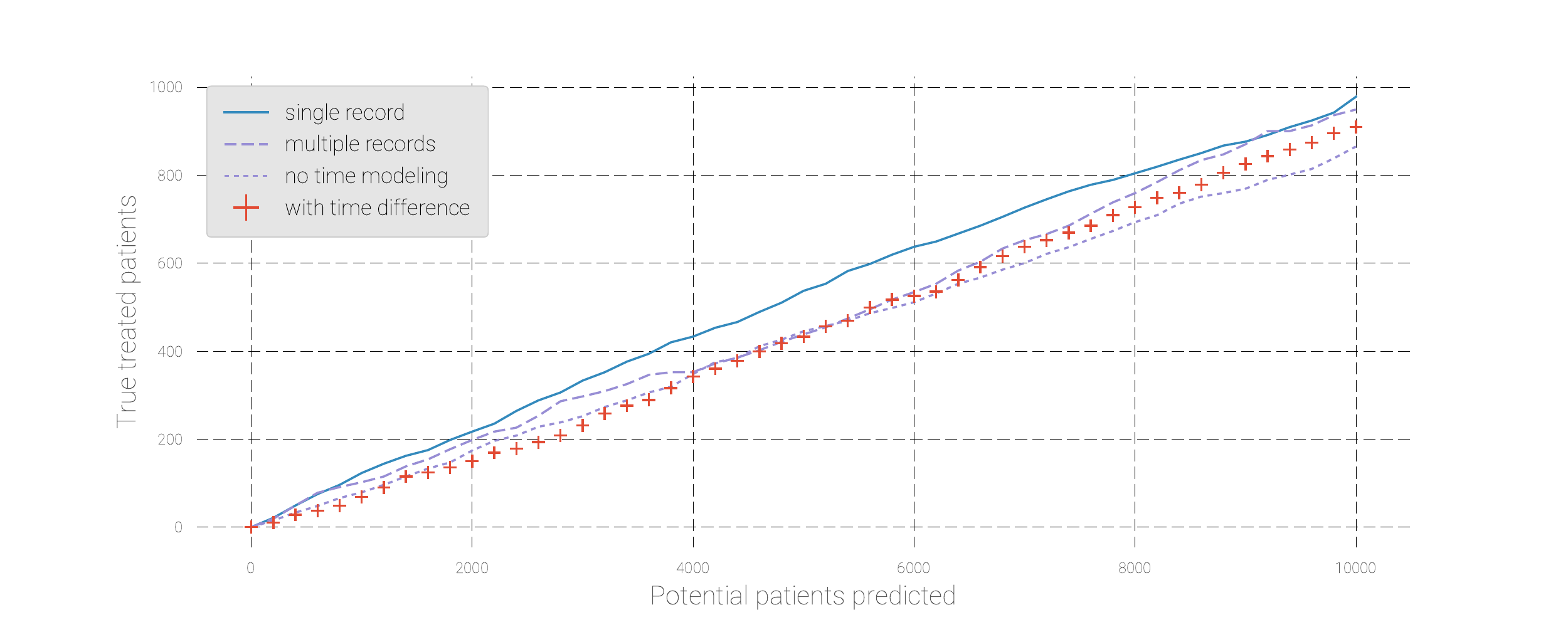}
\end{center}
\caption{The $K$-accuracy curve and the number of successfully predicted patients who received the treatment in the following 12 months. 
\vspace{-7mm}}
\label{fig_accuracy}
\end{figure}

%
\subsection{Proposed Model Framework}

Conventionally, treatment prediction problems have been commonly modeled as a classification problem. For a patient $p$, the model seeks to learn the mapping from the features $\bbx^p$ to the label $y^p$,
\begin{align}\label{eqn_conventional_model}
    \bbx^p = [w_1^p, w_2^p, \dots, w_I^p, z_1^p, z_2^p, \dots, z_D^p] \mapsto y^p.
\end{align}
In the equation, $w_i^p$ denotes the count of the $i$-th service experienced by the patient, where the services may include drugs, other related treatments, examinations, etc., $z_i^p$ denotes demographical information of the patient, 
and $y^p$ is the patient label. 
A machine learning model is then trained upon the features $\bbx^p$ and labels $y^p$ across all $P$ patients using data up to today.

Despite its widespread usage in the healthcare and pharmaceutical industry, such model has 3 major problems: (i) The model is trained to predict who will receive the treatment eventually, but the problem of real interest is who will receive the treatment in the near future. (ii) The services of the patient in the previous months and the services of the patient in the previous years are treated equally, since they are all condensed into the counts $w_i$. (iii) The ordering of different services, e.g. service $i$ before service $j$, are ignored. 

In order to solve these challenges, we introduce a time sensitive model so that the count $w_i$ for each service is separated into the count of services for each day, $w_i = w_{i, 1} + w_{i, 2} + \dots w_{i, T}$ 
%
%
across the $T$ days examined in the data. Also, we could separate the patient label for each day $y = y_1 + y_2 + \dots y_T$, 
%
%
where for patients who received the treatment, i.e. $y = 1$, there is one time index $t$ such that $y_t = 1$ and $y_\tau = 0$ for all other time indexes $1 \leq \tau \leq T, \tau \neq t$, and for patients who have not received the treatment, i.e. $y = 0$, $y_\tau = 0$ for all time indexes $1 \leq \tau \leq T$. The introduction of time tends to make $w_{i, t}$ very sparse, and is therefore likely to result in unstable models. In order to make the generated time-based features more robust, we aggregate service records in $w_{i, t}$ across multiple time indexes $t$ for each the $i$-th service. In specific, for each service $i$ and a pre-chosen {\it index date} $t$ when we examine the label $y_t$, the total amount of services for the previous $\delta$ days is computed and recorded
\begin{align}\label{eqn_aggregate_service}
    l_{i, 1} = \sum_{\nu = t-1}^{t - \delta} w_{i, \nu}.
\end{align}
The $\delta$-day sum is evaluated for $\Delta$ intervals covering $\delta\Delta$ days prior to the index date, i.e. we compute
\begin{align}\label{eqn_aggregate_service_tau}
    l_{i, \tau} = \sum_{\nu = t-(\tau-1)\delta - 1}^{t - \delta\tau} w_{i, \nu}.
\end{align}
for each interval $1 \leq \tau \leq \Delta$. Finally, we train a neural network on the time-dependent features $l_{i, \tau}$ across the $\Delta$ intervals and $I$ services. These time-dependent claim-level features are combined with demographical information to predict the label,
\begin{align}\label{eqn_aggregate_predict}
    [l_{1,1}, l_{2,1}, \dots, l_{1, \Delta}, l_{2,1},l_{2,2},\dots, l_{2, \Delta}, 
    \dots, l_{I,1},l_{I,2},\dots, l_{I, \Delta}, z_1, z_2, \dots z_D] \mapsto y.
\end{align}
As an illustration, Figure \ref{fig_illustration} depicts how claim-level features are constructed with total time indexes $T = 8$, total services $I = 3$, interval length $\delta = 2$, and number of intervals $\Delta = 2$. When $t = 5$, as colored in blue, for each service $i$, we generate a feature $w_{i, 1} + w_{i, 2}$ and another feature $w_{i, 3} + w_{i, 4}$ by aggregating the 4 days prior to $t$ into 2 buckets. 

Thereafter, for positive patients with $y = 1$ who received the treatment, her index date $t$ is set to be the day when the treatment was received. For negative patients with $y = 0$ who have not received the treatment, her index date $t$ is set to be the {\it split date} $t_{\text{split}}$, i.e. the most recent date in the database. 

\section{Experiment}

The dataset has $139,816$ patients and spans the clinical records from 2013 and 2016. We set the split date $t_{\text{split}}$ to be Nov-11-2016, and use only the claim data prior to $t_{\text{split}}$ to generate features, train the model, and perform offline validation. For each of the models trained, it is asked to predict $K$ patients who had not received the treatment prior to or on $t_{\text{split}}$ and whom the model believes are the most likely to receive the treatment. We then examine how many of the predicted $K$ patients indeed received the treatment between Nov-11-2016 and Nov-11-2017, and use the accuracy of prediction as the performance metric. We note that this experiment is analogy to time-series stock prediction, i.e. given all stock performance in the four years prior to Nov-11-2016, train models to predict $K$ stocks which are expected to perform the best in the following year, and then we evaluate how many of these $K$ stocks are indeed the top performer from Nov-11-2016 to Nov-11-2017. We are evaluating the models in a pure {\it online} way, which is much harder than most existing methods which train a model and apply the model to predict the remaining patients in the {\it same} time interval. 

%
\begin{table}[t]
\begin{center}
\label{tab_accuracy}
\scriptsize
\caption{The number of successfully predicted treatments from Nov-11-2016 to Nov-11-2017 among the $K$ selected patients across the 4 models, and the improvement of the proposed model.}
\begin{tabular}{ccccccc} \toprule
$K$ & 1,000 & 2,000 & 3,000 & 5,000 & 7,000 & 10,000 \\\midrule
Proposed Alg. &123	&217	&333&	537	&726	&978	 \\
Count &79	&174	&252	&445	&600	&865	 \\
Count + TD &69	&150&	231	&433	&637&	909	\\\midrule
Proposed Alg. vs.~Count & 55.70\%&	24.71\%&	32.14\%&	20.67\%&	21.00\%&	13.06\% \\
Proposed Alg. vs.~Count + TD &78.26\%&	44.67\%&	44.16\%&	24.02\%&	13.97\%&	7.59\% \\\bottomrule
\end{tabular} 
\vspace{-5mm}
\end{center}
\end{table}

%

%
We construct time-sensitive claim-level features using time indexes $T = 1,411$, services $I = 26$, demographical features $D = 28$ (including $18$ dummy binary variables), interval length $\delta = 91$ (each time bucket is for one quarter), and number of intervals  $\Delta = 2$, probability for negative prior treatment to be included $q = 0.3$. This results in $80$ number of features. The demographical information are constructed using data on the split date. As benchmarks, we consider the conventional model with features in \eqref{eqn_conventional_model} fed into a deep neural network model (named no time modeling in Figure \ref{fig_accuracy}); we also consider adding features that represent the time difference between the diagnosis date and the date when the patient received each of the treatments for the first time (named time difference (TD) in Figure \ref{fig_accuracy}). Out of the $123,260$ potential patients who had not received the treatment by Nov-11-2016, $6,909$ of them ($5.61\%$) received the treatment between Nov-11-2016 and Nov-11-2017. 

A model is trained using the deep neural network with dropout ratio of $20\%$. The deep neural network is better than random forest, xgboost, and regression based models on this expecriment. The $K$-accuracy curve and the number of successfully predicted patients who received the treatment in the following 12 months are presented in Figure \ref{fig_accuracy} and Table \ref{tab_accuracy} respectively. It can be seen that in average, the proposed methods with time-dependent features improve the accuracy by more than $20\%$ in both conventional model with no time information and the model with time difference information. From a business perspective, this indicates that out of 100 sales call, conventional method is expected to identify $8$ to $9$ patients who would receive treatments in the near future and make the deal with them, while the proposed method with time-dependent features is expected to identify $10$ to $11$ patients, resulting in $2$ to $3$ more sales compared to the conventional models. 

%
\section{Conclusion}

This research is developing a time sensitive predictive modeling to first identify patients with high potential to change treatment and most importantly, this model will bring an innovative approach to integrate forecasting technique with predictive modeling to enhance accuracy of the prediction. The overall lift of accuracy rate in forecasting time period doubled the traditional results in the simulation process. More research are undergoing to further improve the accuracy of the final results.

%


%

\medskip

\small

\bibliographystyle{IEEEtran}
\bibliography{Bibliography}

\end{document}